\documentclass{article}

\usepackage{spconf}
\usepackage{graphicx}
\usepackage{array}
\usepackage{epstopdf}
\usepackage{epsfig} 
\usepackage{amsmath}
\usepackage{amsfonts}
\usepackage{amssymb}
\usepackage{cite}
\usepackage{caption}
\usepackage{balance}
\usepackage{subfigure}
\usepackage{multirow}
\usepackage{setspace}
\usepackage{algorithmic}
\usepackage{algorithm}
\usepackage{multirow}
\usepackage{slashbox}

\newcommand{\blist}{\begin{list}\setlength{\topsep 0pt \parsep 0pt}}
\newcommand{\elist}{\end{list}}
\newtheorem{thm}{Theorem}
\newcommand{\bthm}{\begin{thm}\begin{textit}}
\newcommand{\ethm}{\end{textit}\end{thm}}
\newtheorem{dfn}{Definition}
\newcommand{\bdfn}{\begin{dfn}\begin{textit}}
\newcommand{\edfn}{\end{textit}\end{dfn}}
\newtheorem{obn}{Observation}
\newcommand{\bobn}{\begin{obn}\begin{textit}}
\newcommand{\eobn}{\end{textit}\end{obn}}

\newtheorem{lema}{Lemma}
\newcommand{\bdla}{\begin{lema}\begin{textit}}
\newcommand{\edla}{\end{textit}\end{lema}}
\newtheorem{pro}{Property}
\newcommand{\bpro}{\begin{pro}\begin{textit}}
\newcommand{\epro}{\end{textit}\end{pro}}

\newcounter{algnum1}

\newcounter{algnum2}

\newcounter{algnum3}

\newcommand\bit{\begin{itemize}}
\newcommand\eit{\end{itemize}}

\newcommand{\Sec}[1]{Section~\ref{#1}}

\newcommand{\balgo}{
   \begin{description}
   \parskip=0pt
   \topsep=0pt
}
\newcommand{\ealgo}{
   \end{description}
}

\title{Fusing Video and Inertial Sensor Data for Walking Person Identification}
\name{Yuehong Huang, Yu-Chee Tseng
}
\address{Department of Computer Science, National Chiao Tung University, Taiwan\\Emails:\{huang516, yctseng\}@cs.nctu.edu.tw
}
\begin{document}
%
\maketitle
\begin{abstract}
An autonomous computer system (such as a robot) typically needs to identify, locate, and track persons appearing in its sight. However, most solutions have their limitations regarding efficiency, practicability, or environmental constraints. In this paper, we propose an effective and practical system which combines video and inertial sensors for person identification (PID). Persons who do different activities are easy to identify. To show the robustness and potential of our system, we propose a walking person identification (WPID) method to identify persons walking at the same time. By comparing features derived from both video and inertial sensor data, we can associate sensors in smartphones with human objects in videos. Results show that the correctly identified rate of our WPID method can up to 76\% in 2 seconds.
\end{abstract}
\section{Introduction}
\label{sec:intro}
\begin{figure}
	\centering
	\subfigure[]{ 
		\label{fig:scene:ZhongDing01} 
		\begin{minipage}[b]{0.2\textwidth} 
			\centering 
			\includegraphics[width=1.4in]{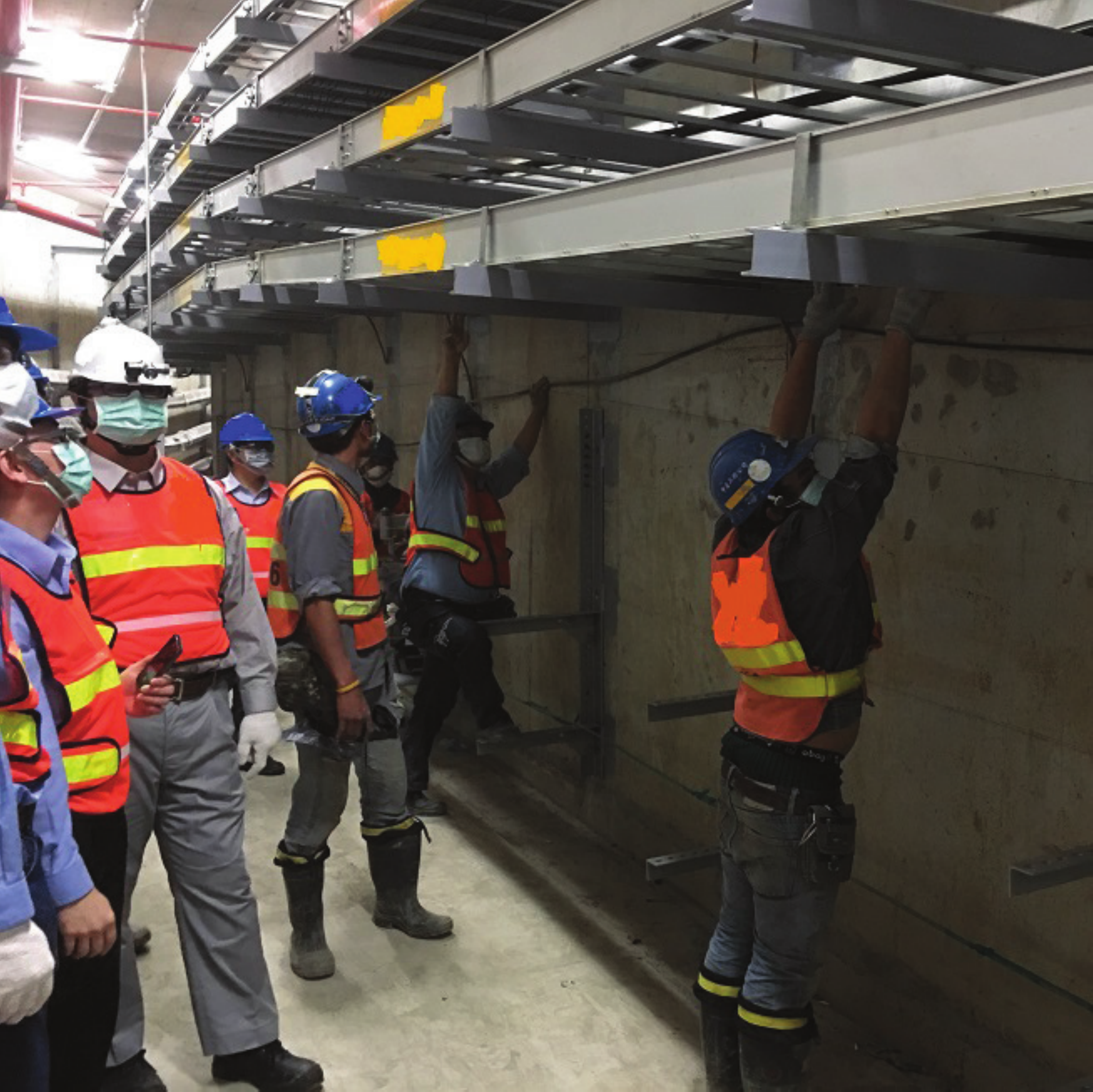} 
	\end{minipage}}%
	\subfigure[]{ 
		\label{fig:scene:labeled} 
		\begin{minipage}[b]{0.2\textwidth} 
			\centering 
			\includegraphics[width=1.4in]{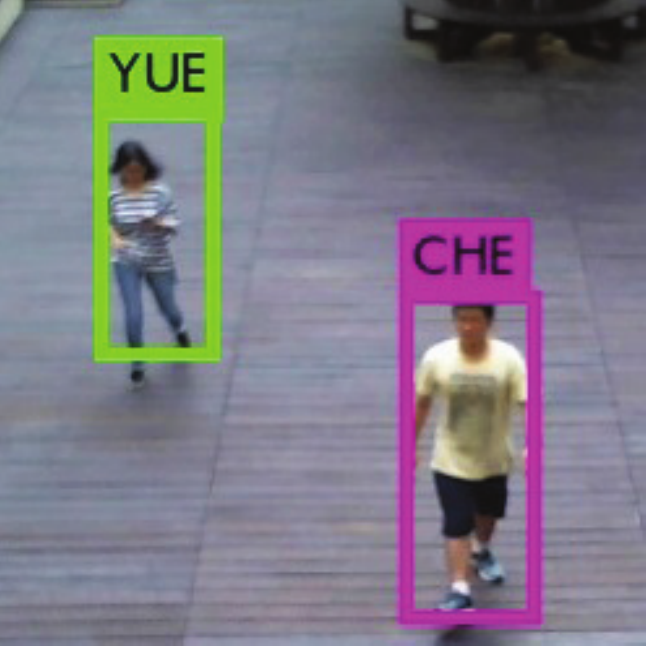} 
	\end{minipage}}  
	
	\caption{Scenes where biological features are difficult to extract.
		\label{fig:scene}}
\end{figure}

Human navigates the world through five senses, including taste, touch, smell, hearing, and sight. We sometimes rely on one sense while sometimes on multiple senses. For computer systems, the optical sensor is perhaps the most essential sensor which captures information like human eyes. Cameras are widely used for public safety and services in hospitals, shopping malls, streets, etc. On the other hand, booming use of other sensors is seen in many IoT applications due to the advances in wireless communications and MEMS. In this work, we like to raise one fundamental question: how can we improve the perceptivity of computer systems by integrating multiple sensors? More specifically, we are interested in fusing video and inertial sensor data to achieve person identification (PID), as is shown in Fig.~\ref{fig:scene:labeled}. 

Efficient PID is the first step toward surveillance, home security, person tracking, no checkout supermarkets, and human-robot conversation. Traditional PID technologies are usually based on capturing biological features like face, voice, tooth, fingerprint, DNA, and iris~\cite{FaceRecognition, tooth, iris}. However, these techniques require intimate information of users, cumbersome registration, training process, and user cooperation. Also, relying on optical sensors implies high environmental dependency (such as lighting, obstacle, resolution, view angle, etc.), thus not suitable for public sites. A scene captured in a construction site is shown in Fig.~\ref{fig:scene:ZhongDing01}, where workers must wear helmets and masks to protect themselves from falling objects and toxic gases. A top view of a courtyard is shown in Fig.~\ref{fig:scene:labeled}. Clearly, recognizing biological features is difficult in such scenarios. Some other recognition approaches are based on wireless signals, but require active participation by users~\cite{WirelessRecognition01, WirelessRecognition02}. The ID-Match method proposed in~\cite{IDMatchRFID} integrates computer vision via depth camera and UHF RFID. It is capable of recognizing individuals walking in groups while wearing RFID tags, thus enabling human-robot interaction. However, this method is handicapped by short range, and all the users need to carry extra RFID tags.

\begin{figure}[t]
	\centering
	\includegraphics[width=8.6cm]{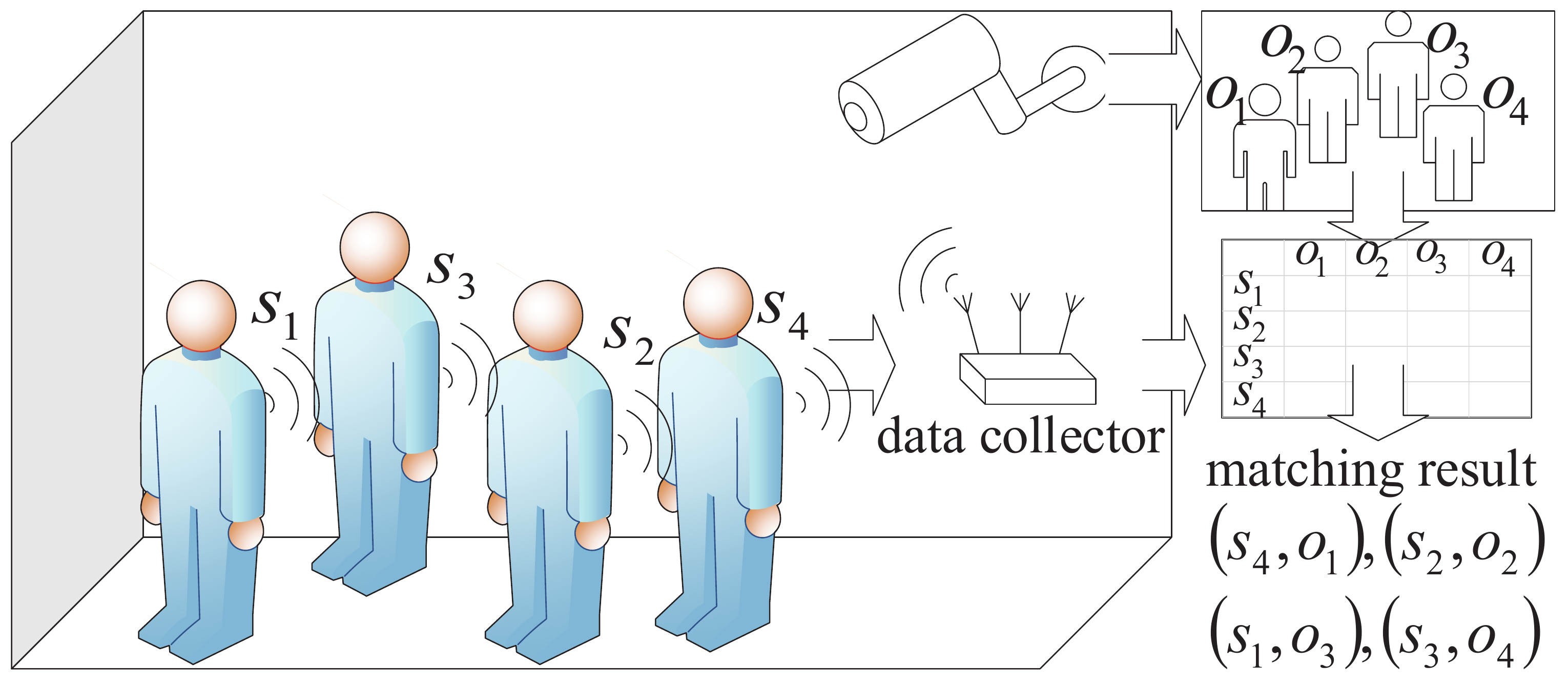}
	\caption{Data fusion workflow.
		\label{fig:IdentificationFlow}}
\end{figure}

In this work, we propose a practical, effective and convenient PID system by combining computer vision and inertial sensor data. Because almost everyone carries a smartphone and almost every smartphone has inertial sensors inside. The main workflow of our PID system is shown in Fig.~\ref{fig:IdentificationFlow}. From video data, a set $O = \{o_{1}, o_{2}, ...\}$ of human objects and their comparable features are retrieved. Similarly, from inertial sensors, a set $S = \{s_{1}, s_{2}, ...\}$ of inertial data and their comparable features are retrieved. Then, the similarity score of each $o_i$ and each $s_j$ is calculated. By analyzing all the similarity scores, the pairing between $O$ and $S$ is derived, which leads to PID result. Inertial sensors are widely used to derive carrier's motions, paths, and physical activities. They are standard modules for current smartphones. On the other hand, we can get motions, traces, and physical activities of people from videos. When persons do different types of activities, it is easy to pair an object with a sensor. But when people do the same activity at the same time, it is difficult to identify persons. So this work only discusses the situation that all the people under camera are walking.

The contributions of this work are as follows. First, we develop a practical, low-cost, and robust PID system. Second, our solution integrates two types of popular sensors. Third, in this work, our matching method focuses on WPID to show the robust of our PID system that combines video and inertial sensor data together.

The rest of this paper is structured as follows. \Sec{sec:method} introduces our PID system and WPID method. Performance evaluation results are presented in \Sec{sec:simu}. Conclusions are drawn in \Sec{sec:conclu}.

\section{PROPOSED WALKING PERSON IDENTIFICATION}
\label{sec:method}

\begin{figure}[t]
	\centering
	\includegraphics[width=7.0cm]{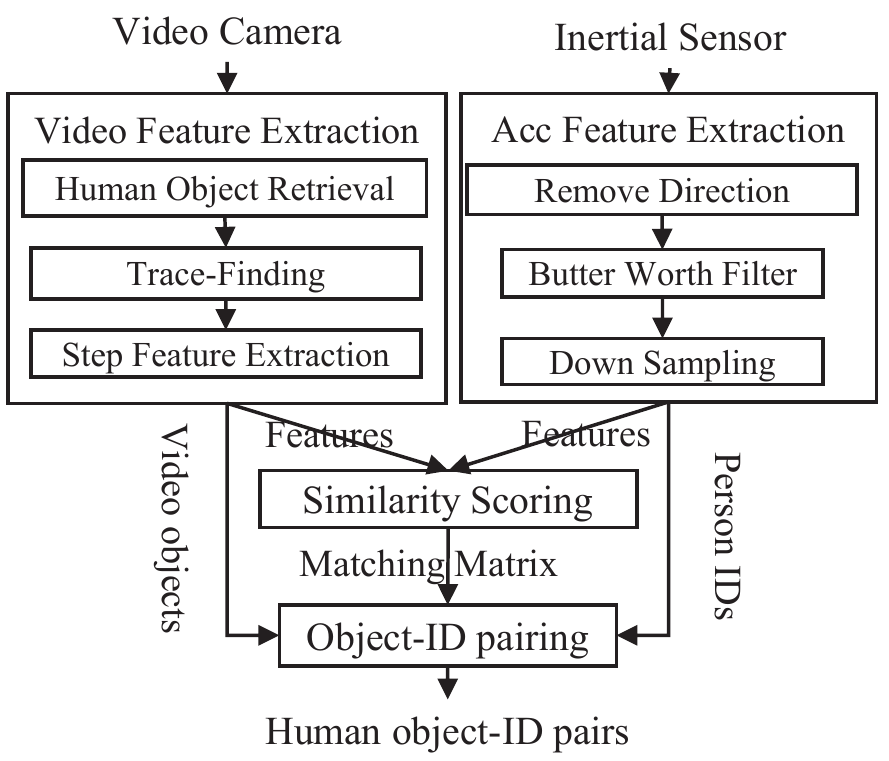}
	\caption{Our PID architecture.
		\label{fig:SystemArchitecture4}}
\end{figure}

We consider an environment in Fig.~\ref{fig:IdentificationFlow} with a video camera and multiple users. The data collected from both camera and smartphones is sent to a server for PID purpose. Our PID system has four software modules as shown in Fig.~\ref{fig:SystemArchitecture4}. The video feature extraction module retrieves human objects and walking traces from a sequence of video frames. The acceleration (Acc) feature extraction module retrieves walking information from acceleration data. The similarity scoring module compares the walking features from both data sources and assigns them similarity scores. The object-ID pairing module couples human objects with smartphones based on the similarity scores. 

\subsection{Video Feature Extraction Module}

\begin{figure}[t]
	\centering
	\includegraphics[width=8.6cm]{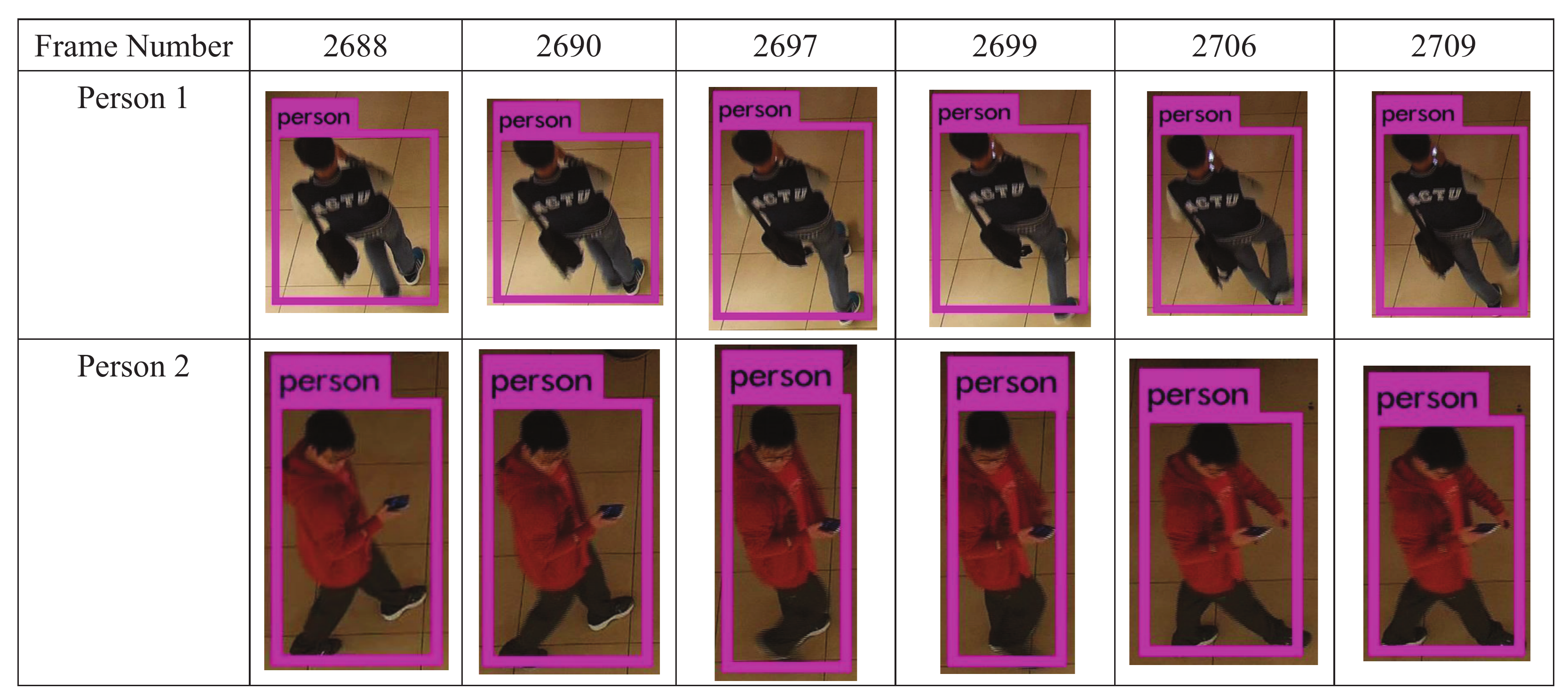}
	\caption{The changes of bounding boxes during walking.
		\label{fig:boxforwalking}}
\end{figure}

The human object retrieval sub-module processes each frame to extract the objects that are recognized as human. It is directly realized by YOLO~\cite{YOLO, YOLO9000, ObjectDetection}. For each frame, YOLO outputs a set $O$ of human objects represented by bounding boxes, and each bounding box is a rectangle inside where YOLO recognizes a human object. The $i$th bounding box of $O$ is denoted by $o_{i}$ and its center, width, and height are denoted by $o_{i}.c$, $o_{i}.w$, and $o_{i}.h$, respectively. Examples are shown in Fig.~\ref{fig:boxforwalking}. 

The trace-finding sub-module is to connect the human objects of adjacent video frames and form continuous traces, where a trace is a sequence of human objects that are regarded as the same person. Efficient object tracking algorithms are available in~\cite{Bewley2016_sort, 7045866, 6909555, Yang, 7410891}, but we design a lightweight tracing method based on movement limitation. Generally, human's running speed is less than $15$ km/h. Assuming a frame rate of $30$ frames per second (fps), in most cases, a person cannot move over $0.1$ of his height between two frames. Based on this assumption, each trace has a \textit{search range} to find its human object in the next frame. The results are some traces connecting human objects in continuous frames.

\begin{figure}[t]
	\centering
	\includegraphics[width=8.6cm]{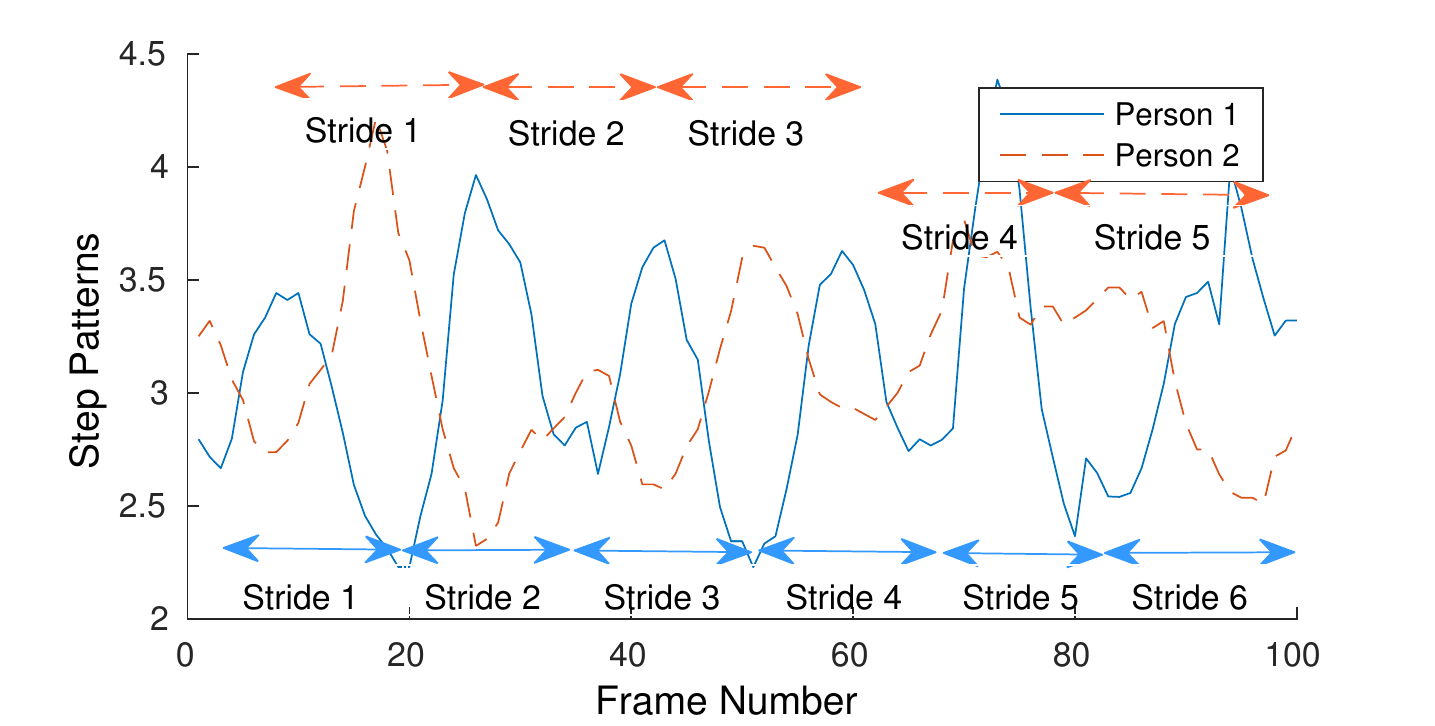}
	\caption{Ratio-features of walking traces.
		\label{fig:steppatternextraction}}
\end{figure}

After trace-finding, the step feature extraction sub-module retrieves walking-related features from each trace. Fig.~\ref{fig:boxforwalking} shows two sequences of frames of two human objects. Suppose our camera has a downward viewing angle. Person 1 walks along a vertical line. When he steps forward, his bounding box becomes taller. When he closes feet, his bounding box becomes shorter. Person 2 walks along a horizontal line. When he steps forward, his bounding box becomes wider. When he closes feet, his bounding box narrows down. As a result, the changes of $o_{i}.h/o_{i}.w$ over time are regarded as step patterns. We use $t_{i}$ to denote the ratio-feature of $i$th trace. Fig.~\ref{fig:steppatternextraction} shows the ratio-features extracted from two persons, who make $6$ and $5$ strides in $100$ frames, respectively. We also mark the ground truth of strides in the graph. As can be seen, the ratio-feature can well present human step patterns.

\subsection{Acc Feature Extraction Module}
In this work, each user carries a smartphone which has installed our application (app), and they can put them in pockets or just hand them. Our software only collects acceleration from the inertial sensor. Since activity recognition from inertial sensor data has been intensively studied, we simply adopt existing solutions~\cite{Senorinmobilephone, sensorbetter}. The sensor data $\hat{a_{i}}$ from the $i$th device is a sequence of acceleration magnitudes after removing direction. Since most energy captured by accelerometer associated with human movements is below 15 Hz~\cite{mathie2003monitoring}, we remove the high-frequency components from $\hat{a_{i}}$. $\hat{a_{i}}$ is low-pass filtered by a $10$th order Butterworth filter with a 15 Hz cut-off frequency~\cite{SensorForStep}. Further, since our frame rate is 30 fps, the simple frequency of $\hat{a_{i}}$ is decreased to 30 per second. After these steps, we get $a_{i}$ as step feature.

\subsection{Similarity Scoring Module}
After retrieving step features from video data and sensor data, we want to answer the following question: How similar is ratio-feature sequence $t_{i}$ to Acc sequence $a_{j}$? The similarity between $t_{i}$ and $a_{j}$ is denoted by $Sim$. In this work, we try to match the extremum positions of two sequences by ignoring their exact values. First, we conduct an extremum detection to find local maximum/minimum points with a window of length $d$. For example, when we set $d=10$, we traverse all the points and their most adjacent $10$ points. If the value of a point is bigger/smaller than all the other $10$ most adjacent points, this point is recorded as a maximum/minimum point. In our experiments, we set $d=10$. A maximum point is marked as $1$, a minimum point is marked as $-1$, and the rest are marked as $0$. This process transforms $t_{i}$ and $a_{j}$ into ternary sequences: $\widetilde{t_{i}}$ and $\widetilde{a_{j}}$. The similarity score between $\widetilde{t_{i}}$ and $\widetilde{a_{j}}$ is defined as:
\begin{equation}
\centering
\label{eq:method2}
Sim(\widetilde{t_{i}},\widetilde{a_{j}}) =
\frac{n}{\sum_{x} dif(\widetilde{t_{i}}[x],\widetilde{a_{j}})}.
\end{equation}
$n$ is the number of extremums in $\widetilde{t_{i}}$ and $dif(\widetilde{t_{i}}[x],\widetilde{a_{j}})$ is defined as:
\begin{equation}
\centering
\label{eq:dif}
dif(\widetilde{t_{i}}[x],\widetilde{a_{j}}) =\left\{
\begin{array}{lr}
0, & \widetilde{t_{i}}[x]=0;\\
|y-x|, & \widetilde{t_{i}}[x] \neq 0~ and~ y~exists;\\
1.5\times d,& otherwise.\\
\end{array}
\right.
\end{equation} 
Here, we scan each binary value $\widetilde{t_{i}}[x]$ of $\widetilde{t_{i}}$. If $\widetilde{t_{i}}[x]=0$, then $dif(\widetilde{t_{i}}[x],\widetilde{a_{j}})$ returns $0$. If $\widetilde{t_{i}}[x] \neq 0$, $dif(\widetilde{t_{i}}[x],\widetilde{a_{j}})$ traverse $\widetilde{a_{j}}$ in the range $x-d$ to $x+d$. $y$, a position in the search range, is the nearest position from $x$ and has $\widetilde{a_{j}}[y] = \widetilde{t_{i}}[x]$. If such $y$ exists in the search range, $dif(\widetilde{t_{i}}[x],\widetilde{a_{j}})$ returns $|y-x|$; if not, $1.5\times d$ is returned. Dividing $n$ by the sum of these differences gives the similarity score between $\widetilde{t_{i}}$ and $\widetilde{a_{j}}$.

\subsection{Object-ID Pairing Module}
After similarity scoring, we get $Sim$. $Sim$ is a two-dimensional array recording all the similarity scores until frame $f$. Let $P_{f}$ be the Object-ID pairing result until frame $f$. The pairing problem is now formulated as a different expression of linear sum assignment problem (LSAP)~\cite{assignment}: 
\begin{equation}
\centering
\label{eq:pairing}
max\sum_{i\in O}{\sum_{j\in S} sim_{ij}p_{ij}},
\end{equation}
$sim_{ij}$ is the similarity score between $i$th human object in $O$ and $j$th sensor in $S$, and the assignment constraints are:
\begin{equation*}
\centering
\label{eq:constraints}
\begin{split}
\sum_{i\in O}p_{ij} \leq 1 & ~~~~~~~~\forall j\in S, \\
\sum_{j\in S}p_{ij} \leq 1 & ~~~~~~~~\forall i\in O, \\
p_{ij}\in \{0,1\} & ~~~~~~~~\forall i\in O, j\in S.
\end{split}
\end{equation*}
We use hungarian algorithm to solve this problem. $p_{ij} = 1$ means that human object $i$ is paired to sensor $j$; $p_{ij} = 0$ means that human object $i$ cannot be paired to sensor $j$. In our work, each frame $f$ can have a pairing result $P_{f}$, and we call the pairing result at this stage as \texttt{Raw Pair} stage result.

However, in practice, the identification result is unstable if we base on our \texttt{Raw Pair} stage result. For example, we may identify one person as Sansa when one frame comes in, but we may identify this person as Jack when next frame comes in, and this person may be identified as Lucy when the frame after the next frame comes in. This problem, which is especially serious when the trace of a person is still short, makes the result rough and hard to see. For this consideration, we propose a \texttt{Refined Pair} stage. In \texttt{Refined Pair} stage, the identification result of a trace not just depends on $P_{f}$, but $P_{1}$ to $P_{f}$. Let $RP_{f}$ be the result generated in the \texttt{Refined Pair} stage for frame $f$. Let $RSim$ be a two-dimensional array, and the value of $rsim_{ij}$ is the number of times that object $i$ has been paired to sensor $j$. The refined pairing problem can be formulated as a LSAP:
\begin{equation}
\centering
\label{eq:rpairing}
max\sum_{i\in O}{\sum_{j\in S}rp_{ij}} \log_{2}{(1+rsim_{ij})},
\end{equation}
subject to:
\begin{equation*}
\centering
\label{eq:rconstraints}
\begin{split}
\sum_{i\in O}rp_{ij} \leq 1 & ~~~~~~~~\forall j\in S, \\
\sum_{j\in S}rp_{ij} \leq 1 & ~~~~~~~~\forall i\in O, \\
rp_{ij}\in \{0,1\} & ~~~~~~~~\forall i\in O, j\in S.
\end{split}
\end{equation*}
Different from $sim_{ij}$, $rsim_{ij}$ is the number of pairing times. $sim_{ij}$ is small, but $rsim_{ij}$ can be very large if the trace of human object $i$ is long. The logarithmic function, shown in Eq.~\ref{eq:rpairing}, is used to weaken the impact of the length of traces on pairing. As $p_{ij}$, $rp_{ij} = 1$ means that human object $i$ is paired to sensor $j$; $rp_{ij} = 0$ means that human object $i$ is not paired to sensor $j$.   

\section{Performance Evaluation}\label{sec:simu}
We have developed a prototype system with one video camera and multiple mobile devices. The camera is Logitech webcam with the resolution of $640\times 480$. To prove that our solution is not device-dependent, we have tried different models of smartphones, including Redmi Note 4X, ASUS ZenFone 3, HTC 10 Evo. The server is a personal computer with an Intel(R) Core(TM) i7-3770 CPU and an NVIDIA GeForce GT 620 graphics card. All devices used in our system are synchronized by the same network time server. We conduct a number of experiments on our WPID method. The average speed of our tracing and WPID method on different pairing stages is around 120 fps. Apparently, our WPID method on two different stages only consumes a few server resources. 

\begin{figure}
	\centering
	\subfigure[]{ 
		\label{fig:variable:3} 
		\begin{minipage}[b]{0.2\textwidth} 
			\centering 
			\includegraphics[width=1.4in]{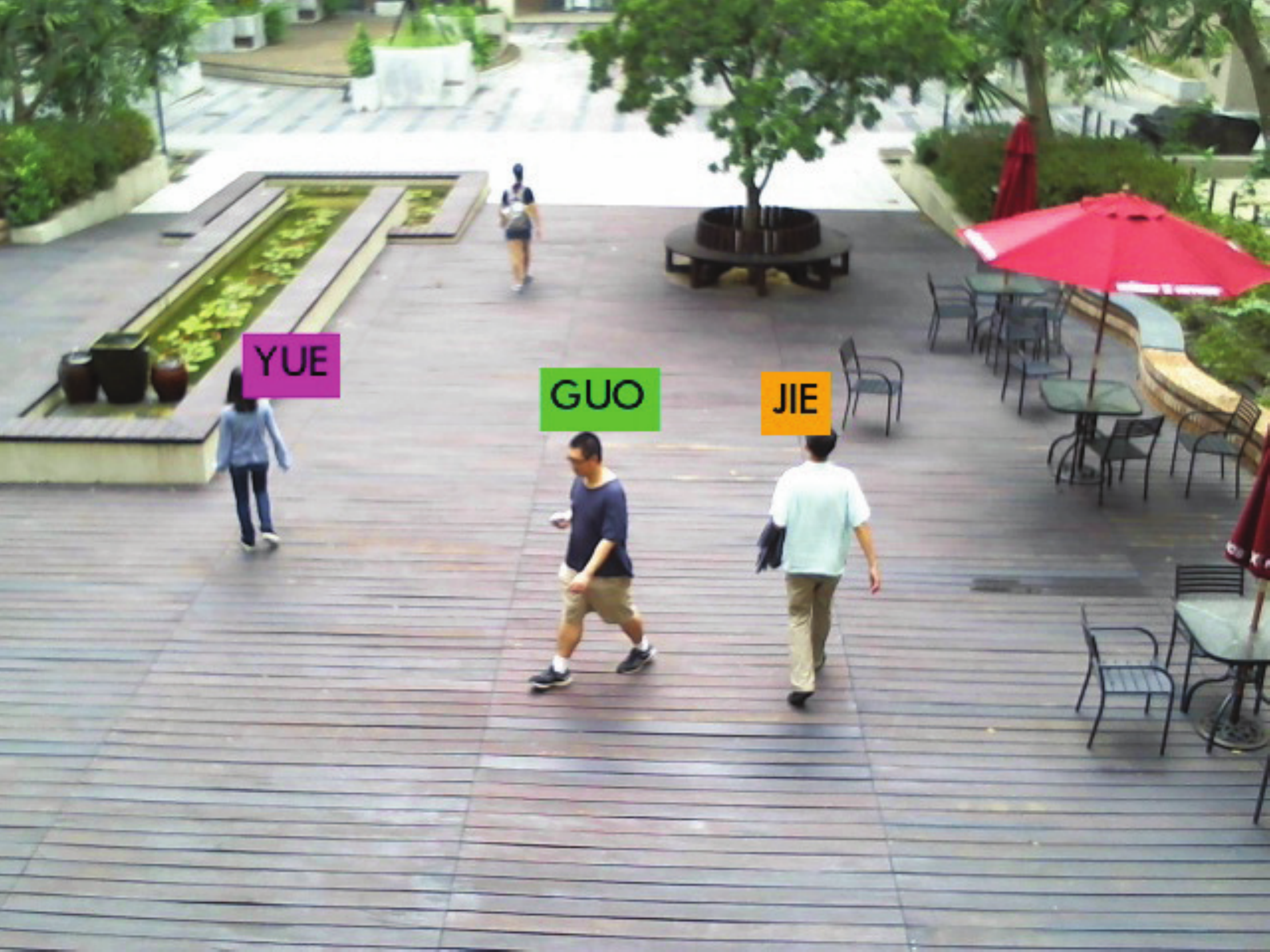} 
		\end{minipage}}%
	\subfigure[]{ 
		\label{fig:variable:4} 
		\begin{minipage}[b]{0.2\textwidth} 
			\centering 
			\includegraphics[width=1.4in]{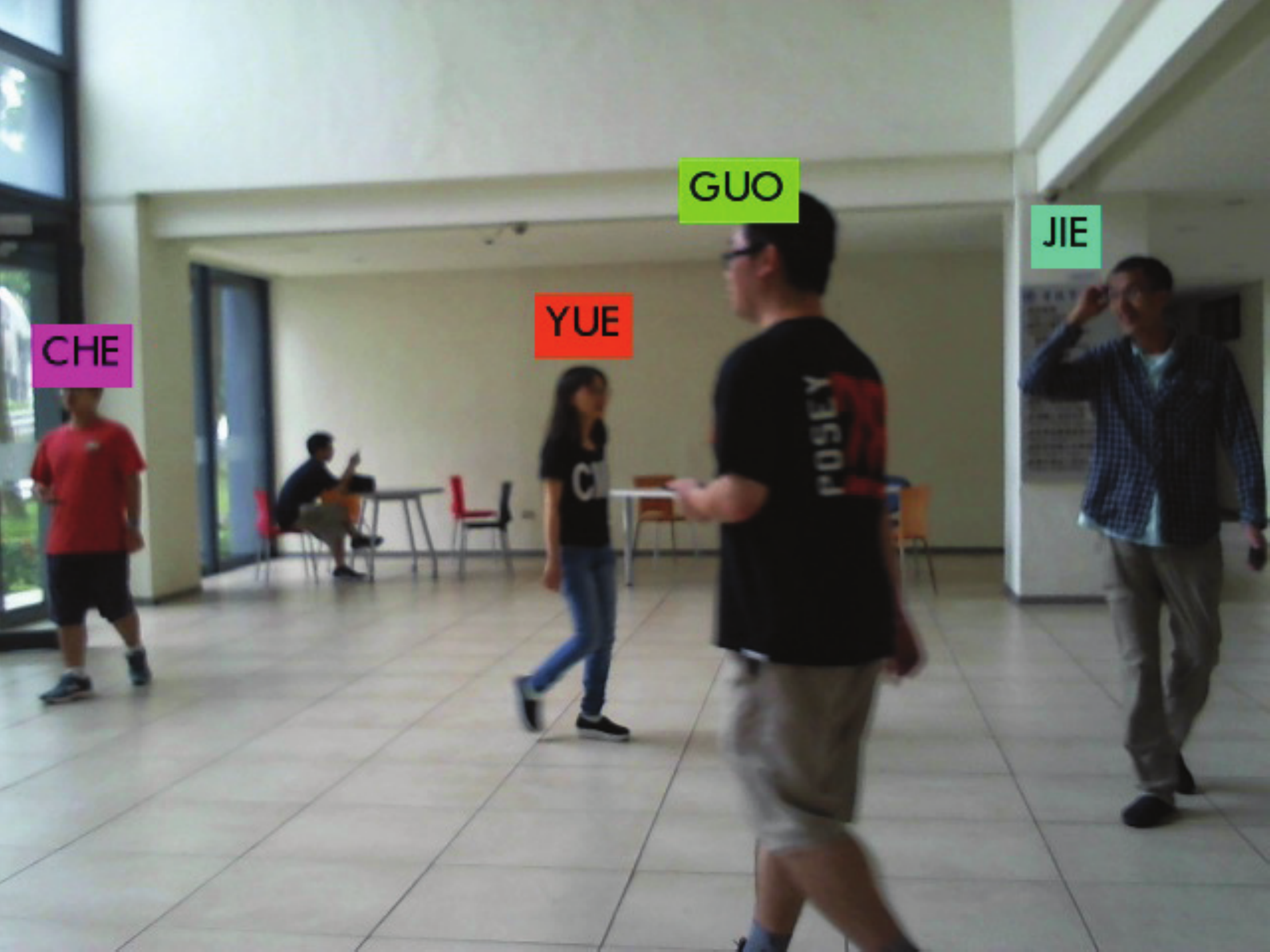} 
		\end{minipage}}
	\caption{Some correctly identified results under different viewing angles and different spaces.
		\label{fig:variable}}
\end{figure}

To show the robustness of our WPID method, experiments are carried out under different areas and viewing angles. A downward viewing angle and outdoor area is set up as shown in Fig.~\ref{fig:variable:3}. The horizontal viewing angle and indoor area is set up as shown in Fig.~\ref{fig:variable:4}. During our experiments, all the persons carry smartphones in their pockets or hands and wander around freely in their styles. As shown in Fig.~\ref{fig:variable}, our WPID method can work under different areas, different view angles, different ways of carrying the smartphones, and different walking styles. The following statistics are all the cases of two persons, and the result of each condition is generated from at least 2000 continuous frames. 

To measure the accuracy of our WPID method, let $O$ be the number of persons having shown in front of the camera until the latest frame. Let $N^{ID}_{i}$ be the number of frames that the $i$th person is identified by our program, and $N^{CD}_{i}$ be the number of frames that the $i$th person is correctly identified by our program. We define our correctly identification rate $R_{cd}$ as: 
\begin{equation}
\centering
\label{eq:ridrate}
R_{cd} = \frac{\sum_{i=1}^{O}N^{CD}_{i}}{\sum_{i=1}^{O}N^{ID}_{i}}.
\end{equation}

Let $TL$ be the length of time that a person is continuously detected by YOLO. If $TL$ is too small, the sequences extracted is too short to be considered as a step pattern. As a result, we set a threshold $TS$ for $TL$. Only when the lengths of two sequences are both larger than $TS$, we do our matching processes. By setting $TS$ from $0.33$ to $4$ seconds, Table~\ref{table:rid} shows the $R_{cd}$ on two stages. From Table~\ref{table:rid}, the increase of $TS$ leads to the increase of $R_{cd}$ in most cases. However, the increase of $R_{cd}$ is not obvious. Also, \texttt{Refined} stage achieves better performance than \texttt{Raw} stage, especially in visual performance.       
\begin{table}[!t]
	\centering
	\caption{The correctly identified rates with different $TS$.}
	\label{table:rid}
	\begin{tabular}{|c|c|c|c|c|c|}\hline
		\backslashbox{Stage}{TS/s} & 0.33\quad & 1\quad & 2\quad & 3\quad & 4 \\
		\hline
		\texttt{Raw} & 0.69 & 0.74 & 0.75 & 0.75 & 0.75\\
		\hline
		\texttt{Refined} & 0.71 & 0.74 & 0.76 & 0.76 & 0.76 \\
		\hline
	\end{tabular}
\end{table}

\section{CONCLUSIONS}\label{sec:conclu}
We propose a new PID system by combining optical and inertial sensors. We design a light tracking algorithm, a WPID method, and two pairing stages. When people do different activities, it is easy to identify persons by comparing behaviors extracted from video and inertial sensor data. So the most complex part of our system is to identify persons who do the same activities at the same time. In this work, we design a WPID method to identify walking persons to show the robustness and potential of our PID system. We conduct extensive experiments and do a lot of discussions to validate the above claims. Results show that the correct identification rate of our WPID method can up to 76\% in 2 seconds. 

\bibliographystyle{IEEEbib}
\bibliography{arxiv_201802_WPID}

\begin{thebibliography}{10}

\bibitem{FaceRecognition}
Y.~Taigman, M.~Yang, M.~Ranzato, and L.~Wolf,
\newblock ``{DeepFace: Closing the Gap to Human-Level Performance in Face
  Verification},''
\newblock in {\em IEEE Conf. on Comput. Vision and Pattern Recognition}, 2014,
  pp. 1701--1708.

\bibitem{tooth}
Dorothy~A. Lunt,
\newblock ``Identification and tooth morphology,''
\newblock {\em J. of the Forensic Sci. Soc.}, vol. 14, pp. 203--207, Apr. 2017.

\bibitem{iris}
F.~Alonso-Fernandez, P.~Tome-Gonzalez, V.~Ruiz-Albacete, and J.~Ortega-Garcia,
\newblock ``Iris recognition based on sift features,''
\newblock in {\em First IEEE Int. Conf. on Biometrics, Identity and Security},
  Sept 2009, pp. 1--8.

\bibitem{WirelessRecognition01}
Mahsan Rofouei, Andrew Wilson, A.J. Brush, and Stewart Tansley,
\newblock ``{Your Phone or Mine?: Fusing Body, Touch and Device Sensing for
  Multi-user Device-display Interaction},''
\newblock in {\em Proc. of the SIGCHI Conf. on Human Factors in Computing
  Syst.}, New York, NY, USA, 2012, CHI '12, pp. 1915--1918, ACM.

\bibitem{WirelessRecognition02}
Sherry Hsi and Holly Fait,
\newblock ``{RFID Enhances Visitors' Museum Experience at the Exploratorium},''
\newblock {\em Commun. ACM}, vol. 48, no. 9, pp. 60--65, Sept. 2005.

\bibitem{IDMatchRFID}
Hanchuan Li, Peijin Zhang, Samer Al~Moubayed, Shwetak~N. Patel, and Alanson~P.
  Sample,
\newblock ``{ID-Match: A Hybrid Computer Vision and RFID System for Recognizing
  Individuals in Groups},''
\newblock in {\em CHI Conf. on Extended Abstracts on Human Factors in Comput.
  Syst.} 2016, pp. 7--7, ACM.

\bibitem{YOLO}
Joseph Redmon, Santosh Divvala, Ross Girshick, and Ali Farhadi,
\newblock ``You only look once: Unified, real-time object detection,''
\newblock in {\em IEEE Conf. on Comput. Vision and Pattern Recognition (CVPR)},
  2016.

\bibitem{YOLO9000}
Joseph Redmon and Ali Farhadi,
\newblock ``{YOLO9000: Better, Faster, Stronger},''
\newblock in {\em IEEE Conf. on Comput. Vision and Pattern Recognition (CVPR)},
  2017.

\bibitem{ObjectDetection}
Ross Girshick, Jeff Donahue, Trevor Darrell, and Jitendra Malik,
\newblock ``{Rich Feature Hierarchies for Accurate Object Detection and
  Semantic Segmentation},''
\newblock in {\em IEEE Conf. on Comput. Vision and Pattern Recognition (CVPR)},
  Washington, DC, USA, 2014, pp. 580--587.

\bibitem{Bewley2016_sort}
Alex Bewley, Zongyuan Ge, Lionel Ott, Fabio Ramos, and Ben Upcroft,
\newblock ``Simple online and realtime tracking,''
\newblock in {\em IEEE Int. Conf. on Image Process.}, 2016, pp. 3464--3468.

\bibitem{7045866}
J.~H. Yoon, M.~H. Yang, J.~Lim, and K.~J. Yoon,
\newblock ``Bayesian multi-object tracking using motion context from multiple
  objects,''
\newblock in {\em IEEE Winter Conf. on Applicat. of Comput. Vision}, Jan 2015,
  pp. 33--40.

\bibitem{6909555}
S.~H. Bae and K.~J. Yoon,
\newblock ``Robust online multi-object tracking based on tracklet confidence
  and online discriminative appearance learning,''
\newblock in {\em IEEE Conf. on Comput. Vision and Pattern Recognition}, June
  2014, pp. 1218--1225.

\bibitem{Yang}
Min Yang and Yunde Jia,
\newblock ``Temporal dynamic appearance modeling for online multi-person
  tracking,''
\newblock {\em Comput. Vis. Image Underst.}, vol. 153, no. C, pp. 16--28, Dec.
  2016.

\bibitem{7410891}
Y.~Xiang, A.~Alahi, and S.~Savarese,
\newblock ``Learning to track: Online multi-object tracking by decision
  making,''
\newblock in {\em IEEE Int. Conf. on Comput. Vision (ICCV)}, Dec 2015, pp.
  4705--4713.

\bibitem{Senorinmobilephone}
Shane Colton,
\newblock ``The balance filter: a simple solution for integrating accelerometer
  and gyroscope measurements for a balancing platform,'' 2007.

\bibitem{sensorbetter}
Muhammad Shoaib, Stephan Bosch, Ozlem~Durmaz Incel, Hans Scholten, and Paul
  J.~M. Havinga,
\newblock ``Fusion of smartphone motion sensors for physical activity
  recognition,''
\newblock in {\em Sensors}, 2014, vol.~14, pp. 10146--10176.

\bibitem{mathie2003monitoring}
M.~Mathie,
\newblock {\em Monitoring and Interpreting Human Movement Patterns Using a
  Triaxial Accelerometer},
\newblock University of New South Wales, 2003.

\bibitem{SensorForStep}
Melania Susi, Valérie Renaudin, and Gérard Lachapelle,
\newblock ``{Motion Mode Recognition and Step Detection Algorithms for Mobile
  Phone Users},''
\newblock {\em Sensors}, vol. 13, no. 2, pp. 1539--1562, 2013.

\bibitem{assignment}
Rainer~E. Burkard and Eranda {\c{C}}ela,
\newblock {\em Linear Assignment Problems and Extensions}, pp. 75--149,
\newblock Springer US, Boston, MA, 1999.

\end{thebibliography}

\end{document}